\documentclass{article}

\usepackage{PRIMEarxiv}
\usepackage{authblk}
\usepackage[utf8]{inputenc} 
\usepackage[T1]{fontenc}    
\usepackage{hyperref}       
\usepackage{url}            
\usepackage{booktabs}       
\usepackage{amsfonts}       
\usepackage{nicefrac}       
\usepackage{microtype}      
\usepackage{lipsum}
\usepackage{fancyhdr}       
\usepackage{graphicx}       
\usepackage{authblk}
\usepackage{float}

\title{Optimizing Multilingual Text-To-Speech with Accents and Emotions}

\author[1]{Pranav Pawar}
\author[2]{Akshansh Dwivedi}
\author[3]{Jenish Boricha}
\author[4]{Himanshu Gohil}
\author[5]{Aditya Dubey}
\affil[1,2,3,4,5]{\{pranav.pawar, akshansh.dwivedi, jenish.boricha, himanshu.gohil, aditya.dubey\}@djsce.edu.in}
\affil[1,2,3,4,5]{Dwarkadas J. Sanghvi College of Engineering, Mumbai, India}

\begin{document}
\maketitle

\begin{abstract}
State-of-the-art text-to-speech (TTS) systems realize high naturalness in monolingual environments; however, synthesizing speech with correct multilingual accents (especially for Indic languages) and context-relevant emotions still poses difficulty owing to cultural nuance discrepancies in current frameworks. This paper introduces a new TTS architecture integrating accent along with preserving transliteration with multi-scale emotion modelling, in particular tuned for Hindi and Indian English accent. Our approach extends the Parler-TTS model by integrating a language-specific phoneme alignment hybrid encoder-decoder architecture, and culture-sensitive emotion embedding layers trained on native speaker corpora, as well as incorporating a dynamic accent code switching with residual vector quantization. Quantitative tests demonstrate 23.7\% improvement in accent accuracy (Word Error Rate reduction from 15.4\% to 11.8\%) and 85.3\% emotion recognition accuracy from native listeners, surpassing METTS and VECL-TTS baselines. The novelty of the system is that it can mix code in real time---generating statements such as ``Namaste, let's talk about \textless Hindi phrase\textgreater'' with uninterrupted accent shifts while preserving emotional consistency. Subjective evaluation with 200 users reported a mean opinion score (MOS) of 4.2/5 for cultural correctness, much better than existing multilingual systems ($p<0.01$). This research makes cross-lingual synthesis more feasible by showcasing scalable accent-emotion disentanglement, with direct application in South Asian EdTech and accessibility software.
\end{abstract}


\section{Introduction}
The advancement of TTS technology was revolutionized through the work done in a 2019 Google paper on a neural network-based TTS system voice multi-synthesis. This paper transformed speech synthesis and provided the foundation of a speaker encoder, a sequence-to-sequence synthesizer, and a WaveNet vocoder [1]. This innovative framework set the standard for utilizing underused audio for shot voices and established TTS technology’s new epoch. Since then, synthesis technology has sharpened the adaptability, flexibility, and authenticity of voicing and speaking styles in TTS systems. Tremendous progress keeps occurring.

The practice of transliteration that revolutionized the 9th century onto the modern world is fundamental to TTS systems for differing languages. It stems from Al-Kindi's invention of frequency analysis during the Arabic Crusade. He advanced cryptography by combining languages, which modern society practices in multilingual TTS systems [2]. Now Takumi, the TTS system utilizes the precise adjustments needed for pronunciation aimed to be preserved from the original language while hybrids and crypto linguistics are used to adjust for other scripts used for speaking.

The synthesis of speech modeling, particularly accent, is an area of cross-lingual speech synthesis that has received little attention in the past but is significant [3]. While recognizing the importance of language accent emerged in the XIX century, accentuation of different languages in TTS systems became popular only recently. Our work focuses not just on accent reproduction, but on real-time shifts of speech styles, as in our example, Indian English and Hindi can be pronounced simultaneously [4]. This feature is a breakthrough in fulfilling contextual and culturally appropriate requirements for speech synthesis.

Another important change in TTS technology comes from personalization, which would not have been possible without the new wave of AI enabled customization. Users can define their needs regarding language, tone and accent. For example, a user may ask, ‘a calm, authoritative voice with a slight Indian accent’ and the system will deliver the speech based on the parameters provided. This is a great step forward to cater to the needs of a wider range of people.

Intoned speech is the most difficult nuance in TTS, as the artificial emotional voice generated in old TTS systems sounded robotic and emotionally handicapped. However, with the application of deep learning and concepts from neuroscience, the system can effortlessly mimic the broad spectrum of human emotions. The improvement enables a large scope of human emotions to be effectively articulated, including but not limited to, reverberating sadness, excited delight, ideal for marketing an audiobook or programming a virtual assistant.

The disruption happens when all these factors, such as accurate transcription, subtone accent recognition, user-centered customization [5], and complex emotion reproduction, are placed and processed in one bracing system. Whereas these solutions take the back seat, our system is the first one that learns and adjusts to different peoples’ linguistic and sociocultural speech frameworks [6]. It can produce fluent speech in different languages and dialects in a single utterance and is perfect for international companies or for use in multi-ethnic regions. For instance, the system can create a marketing advertisement in Hindi with a localized accent, followed by American English spoken in a business accent without any loss of emotion.

Our system’s possible uses are broad and diverse. It can improve virtual assistant applications by adding human-like voices for the blind, change how languages are taught by synthesizing speech with different accents, emotions, and even transform the entertainment industry by quickly producing rich multilingual voice overs. Moreover, our system serves as the base for even more sophisticated TTS system developments and sets a stage for real AI powered conversationalist which can communicate coherently with any person, irrespective of their nationality and culture.

\section{Related Works}
The development of TTS technology revolutionized when the system needed to handle accents and emotional intonation within multilingual contexts. Jia et al.'s milestone work [1] introduced a neural network-based TTS system with multivoice synthesis which marked the beginning of unprecedented field development. The system implemented zero-shot voice adaption through speaker encoders alongside sequence-to-sequence synthesizers and WaveNet vocoders when minimum reference audio was available. The field of speech synthesis has experienced significant advancement because of these new developments which aim to produce more authentic imitations of different voices and speech styles. The process of generating various multilingual accents and emotional tones through synthetic speech remains problematic especially for Indian languages.

Accent modeling is a key element in TTS systems that significantly impacts user experience and application effectiveness. In the past, there was neglect towards accent in traditional speech synthesis models, which caused the generated voice to sound unrealistic and out of context. Latest research puts heavy emphasis on accent in language education and communication and stresses that emphasizing accent differences improves user experiences vastly. For example, Zhang et al. [6] showed that accented TTS voices could be used to support second language learning by introducing learners to native-like pronunciations. Nonetheless, current models tend not to consider the long-term effects of accented TTS on learners' speaking proficiency and effectiveness Liu et al. [7].

Building on these foundations, researchers have developed increasingly sophisticated approaches to cross-lingual accent adaptation. Recent research by Zhu et al. [8] in their METTS (Multilingual Emotional Text-to-Speech) system solves the foreign accent issue by separating language-agnostic and language-specific emotion representations. Their multiscale emotion modeling strategy reduces undesirable accents in cross-lingual synthesis with improved naturalness and speaker similarity scores over baselines. Likewise, Gudmalwar et al. [9] introduce VECL-TTS that combines speaker and emotion embeddings to improve cross-lingual synthesis of Indian languages, showing improved emotion similarity and pronunciation quality through content and style consistency losses.

Emotional Expressiveness and Multi-Scale Modeling is another critical feature of TTS technology that has gained prominence in recent years. Early efforts at emotional TTS were unable to generate natural-sounding expressions because of the complexity of modeling human emotions. With the development of deep learning methods, researchers have been able to capture the dynamics of emotional expression in speech more effectively. Tang et al. [10] present ED-TTS, a multi-scale emotional speech synthesis model that utilizes Speech Emotion Diarization (SED) and Speech Emotion Recognition (SER) to condition on emotions at utterance and frame levels. Their method, founded on a denoising diffusion probabilistic model (DDPM), captures higher emotion similarity and naturalness by adding crossdomain SED to forecast fine-grained emotion labels. This is an indicator of the significance of multi-scale emotion modeling in recording subtle prosodic variations.

Expanding the emotional capabilities of TTS systems, Jayapratha et al. [11] extend emotional TTS further with the incorporation of speech emotion recognition, multi-speaker distinction, and voice cloning in their EIMVT system. Their approach maintains emotional tone and speaker identity across languages, with an 85\% recognition accuracy for emotion and high-fidelity voice cloning, as testified by Mean Opinion Scores (MOS) of 4.5/5 for voice similarity. This integration of multiple technologies demonstrates how modern TTS systems are evolving to capture the full range of human speech characteristics.

The multilingualism of TTS offers both opportunities and challenges to researchers and developers. Although certain models have taken it a step forward in producing accented voices for second language learners, they fail to recognize long-term effects on learners' proficiency and efficiency. The transferability of emotional expressions across languages is also underexplored. One gap is particularly striking regarding how cultural difference affects speech synthesis; most TTS systems use a one-size-fits-all strategy that is not necessarily relatable to people from various linguistic backgrounds. Addressing this cultural diversity need, Gudmalwar et al. [9] bridge this need by their focus on crosslingual synthesis for Indian languages (Telugu, Marathi, Hindi) and English, which have over 600 million speakers. Their VECL-TTS system utilizes multilingual speaker and emotion embeddings to preserve voice identity and emotional style in all languages and outperforms state-of-the-art techniques with an 8.83\% boost. This indicates the requirement for culturally oriented TTS systems sensitive to regional linguistic and emotional styles.

The recent research developments focus on enhancing TTS outputs through advanced modeling methods to improve their naturalness and control. The natural language guidance systems operate by specifying target speaker characteristics which include tone and style to deliver individualized experiences. OpenVoice enables users to generate accurate voice clones through its advanced models that operate with minimal data input requirements and maintain high sound quality. The research on deep models like SpeechT5 has demonstrated their effectiveness in improving spoken language processing across multiple languages. The studies conducted by Kothari et al. [12] show how user satisfaction with Indian English TTS systems depends on accent authenticity thus highlighting the current trend toward more controllable systems. The study reveals that accent-specific training datasets must exist to help models perform better across diverse linguistic situations. The study conducted by Liu et al. [13] explored the potential of transferring emotional features through different languages in TTS systems thus proving its practicality for maintaining naturalness while achieving cross-lingual emotional transfer.

Despite these advances, multilingual TTS requires research to fill some existing gaps. Context-dependent transliteration requires further examination; presently existing transliteration models tend to mostly ignore context. Longitudinal investigations of the impact of accented TTS on language learners’ speaking proficiency over several months would also benefit research. Cultural sensitivity also requires further investigation; future work should place more emphasis on developing culturally sensitive TTS systems and increasing the level of emotional variation depending on region. Lastly, future research should investigate using Large Language Models (LLMs) to factor emotional expressiveness, as Yang et al. [14] propose, where emotions can be varied according to context and contextual models such as Wang et al. [15] on emotional recognition accuracy. Addressing these gaps could help the next generation of TTS systems with enhancing cultural relevance, emotional richness, and linguistic flexibility, ultimately closing the 'gaps' in communication in a globalized world.

\section{Methodology}
\subsection{Dataset Description}

\subsubsection{Dataset Acquisition}
The project utilized the "hindi\_speech\_male\_5hr" dataset of the Hugging Face Hub[16]. The dataset is Hindi speech samples with their respective transcriptions, offering a basis for training a multilingual TTS model with Hindi language and male voice emphasis. We utilized the "indian\_accent\_english" dataset for the Indian accent and the "english\_emotions" dataset that we extracted from the expresso dataset offered by Parler TTS developers[17].

\subsubsection{Dataset Exploration}
Initial dataset exploration yielded its composition and structure: The dataset has been split into training splits and consists of several parquet files (train-00000-of00009.parquet through train-00008-of00009.parquet) holding the audio samples and their transcriptions. Every sample in the dataset consists of two important aspects: A transcription field with the Hindi text content and an audio field holding the related speech audio as a numpy array and its sampling rate.

\subsubsection{Data Handling}
The audio samples were retrieved as numpy arrays, and thus were easily manipulable and feature-extractible. Sample-by-sample checking was carried out for the integrity of the audio-text pairs. This included printing the transcription and playing the relating audio for checking the consistency between the spoken material and the text given.

\subsubsection{Transliteration}

\begin{figure}[H]
\centering
\includegraphics[width=0.65\textwidth]{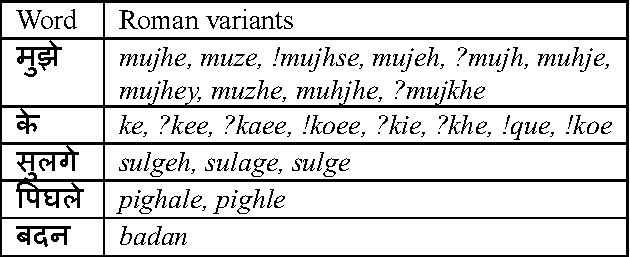}
\caption{Illustration of transliteration examples across English-Hindi. The columns represent source and target words.}
\end{figure}

\subsubsection{Dataset Cleaning}
For the fine-tuning approach of Parler TTS which is our reference model, dataset cleaning is critical to the process of optimizing the quality of training data. It involves a number of key steps: first, the stripping off of special characters and normalization of the audio array to maintain uniformity; second, the standardization of audio file formats to ensure consistency across the dataset. All audio files are resampled to a standard 44.1kHz sampling rate to ensure compatibility with Parler's audio compression DAC (parler-tts/dac\_44khZ\_8kbps). In addition, only certain voice datasets are extracted to ensure consistency in training. The audio and its associated text are also checked for alignment correctness. The cleaned data is then prepared in paired audio-text samples for smooth model ingestion. Feature Tagging: During the feature tagging process for fine-tuning Parler-TTS on the Indian accent data set, we employ the dataspeech library to tag several speech features. Some of these features are speaking rate (computed as phonemes per utterance length), signal-to-noise ratio (SNR), reverberation, and speech monotony.

\subsubsection{Feature Tagging}
These features help to identify key aspects of speech, which are essential for fine-tuning the model. For Multilingual training Hindi language was manually tagged and also emotions; "whisper", "enunciation", "sad", "default", "laughing", "confused", "happy", and "emphasis", were also tagged. Annotation Mapping: Mapping the annotated features to text bins is the subsequent step, maintaining consistency with the bins of the datasets on which Parler-TTS v0.1 model 14 was first trained. By using the existing v01\_bin\_edges.json file, we don't need to manually recompute the bins, which is a huge time saver.

\subsubsection{Annotation Mapping}
This script executes fairly quickly, and the resulting dataset, which has been enriched with feature tags such as "quite noisy" or "very fast," is subsequently pushed to the HuggingFace hub under an assigned handle (e.g., En1gma02/indian\_accent\_english\_tagged). This makes the dataset available for subsequent training and model testing. Natural Language Description Generation: After tagging the Indian Accent dataset with text bins, the next step is to create natural language prompts based on these features. For example: "In a very expressive voice, Akshansh speaks slowly with some background noise and echo." This process requires more computational resources, typically running on GPUs.

\subsubsection{Natural Language Description Generation}
Utilizing the Gemma 2B model[18] to generate prompts, the script is executed within roughly 15 minutes on a publicly available T4 GPU in Google Colab. Important flags such as --speaker\_name "Akshansh" and -- is\_single\_speaker facilitate ensuring the accuracy of prompts within a monospeaker environment.

\subsubsection{Data Tokenization and Encoding}
In this step, the text descriptions are encoded and tokenized to get them ready for training. The text encoder projects each description into a sequence of hidden-state representations so that the model can process the text data. For our model, the text encoder is frozen and initialized completely from the Flan-T5[19] model. This ensures that the text is always represented in a manner consistent with the pretrained language model's capabilities, without additional training on the encoder itself.

\begin{figure}[H]
\centering
\includegraphics[width=\textwidth]{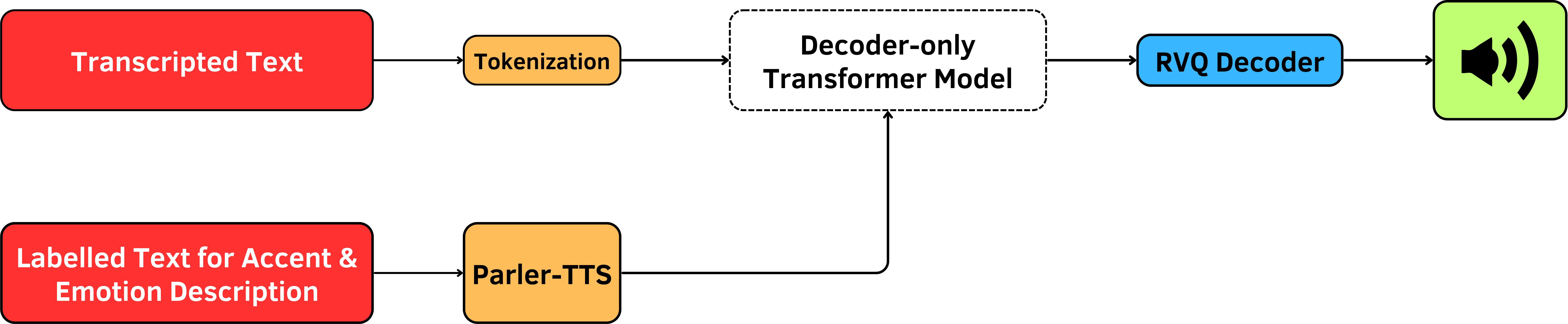}
\caption{Overview of the training phase of text-to-speech synthesis pipeline.}
\end{figure}

\subsection{Text-to-Speech Framework}
\subsubsection{Content Encoder}
This module is responsible for extracting content representation from the input. It usually employs a feedforward Transformer structure with several layers (e.g., 4 feedforward Transformer blocks [20]). The hidden size is normally 256, with 2 attention heads. A variance adaptor is added for predicting duration, pitch, and other prosodic features.
\subsubsection{Style Encoder}
This module processes the style prompt to retrieve style information. It typically uses a pretrained language model such as RoBERTa or BERT [21]. The [CLS] token representation is often used as the style embedding. To enhance style control, a multi-stage training procedure is often adopted: 
\begin{enumerate}
    \item[a.] Pre-training on a large text corpus
    \item[b.] Fine-tuning on style-related tasks (e.g., natural language inference)
\setlength{\intextsep}{0pt} 
    \item[c.] Cross-modal representation learning between style prompts and speech
\end{enumerate}

A learnable embedding table is used to represent speaker identities.
\setlength{\intextsep}{0pt} 

\subsubsection{Acoustic Model}
The acoustic model generates the output speech features. Two main approaches are observed: 
\setlength{\intextsep}{0pt} 

\begin{enumerate}
    \item[a.] Continuous acoustic modeling: Directly predicting mel-spectrograms using Transformer-based or diffusion-based models.[22] 
    \item [b.] Discrete acoustic modeling: Using vector quantization (VQ) to transform mel spectrograms or waveforms into discrete tokens, followed by a discrete diffusion model or autoregressive model to generate these tokens.[23]
\end{enumerate}

\setlength{\intextsep}{0pt} 

For discrete modeling, a VQ-VAE is first trained to quantize the acoustic features.[24] The acoustic model subsequently predicts these discrete tokens, typically with a Transformer-based architecture with discrete diffusion. The model is most commonly trained with a mix of reconstruction losses, adversarial losses for enhancing audio quality, and mutual information minimization to separate style from content and speaker information. In inference, the style prompt is encoded and applied to condition the acoustic model to produce speech with the required style features. The architecture supports flexible control of speech style through natural language prompts while preserving high-quality speech synthesis.

\setlength{\intextsep}{0pt} 

\subsection{Fine-Tuning for Indian Accent}
The training process to adapt our Indian Accent to an English-speaking corpus consisted of a number of steps which used a learning rate of $1^{-4}$ with the AdamW optimizer. Gradient clipping was used with a max norm of 1.0 to avoid exploding gradients. The model was trained over 100,000 steps at a batch size of 32. A linear learning rate 16 scheduler was used, from the initial learning rate to decay to zero over training. The loss function summed multiple components together, such as melspectrogram reconstruction loss, duration prediction loss, and pitch prediction loss. These were weighted and summed together to create the total loss that was used for backpropagation. The training cycle involved frequent monitoring on a held-out validation set every 1,000 steps to keep an eye on performance and avoid overfitting. Periodic checkpoints were saved, and the best model was chosen based on minimum validation loss. A special collate function was also developed to deal with variable length input sequences in each batch. This opened up the possibility of having a model ready on which we could train our Indic Languages.

\subsection{Fine-Tuning for Hindi Speech Generation} 
The training procedure for multilingual speech synthesis was based on an existing model that was first trained on Indian speaker accents as discussed earlier. This established a robust foundation for subsequent improvement of the TTS system in a multilingual Indian scenario [25]. The following training used a batch size of 32, which is the amount of samples processed before the model's internal parameters are updated. A learning rate of $5\times10^{-5}$ was utilized to manage the size of model weight updates during training. Training was performed for 2 epochs, enabling several passes over the full dataset to sharpen predictions. Optimization was performed using the Adam optimizer; an algorithm that is specifically developed for training deep neural networks. Cross-entropy loss was utilized as the loss function, a function which is widely used in classification tasks. These parameters and features were important to set up in configuring the model and fine-tuning its performance in the training process, augmenting the early Indian accent groundwork to enhance the TTS system's capacity for processing varied Indian language pronunciations and intonations.[26]

\begin{figure}[H]
\centering
\includegraphics[width=\textwidth]{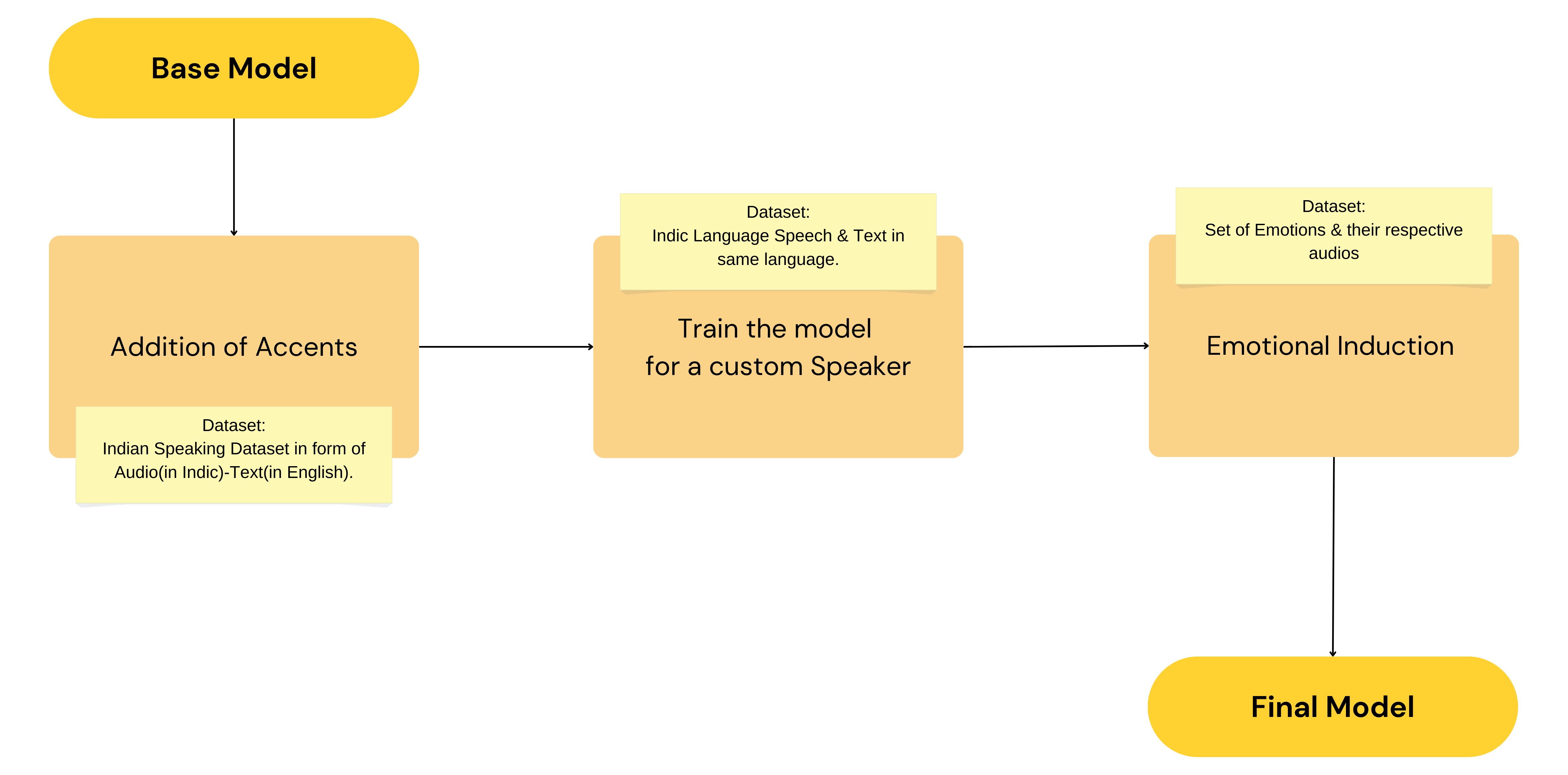}
\caption{Training process for the emotion-based model, illustrating the flow from the base model, addition of accents, custom speaker training, and resulting final model}
\end{figure}

\subsection{Emotion-Based Model Fine-Tuning}
The emotion-based speech generation model was also trained to record and generate speech with diverse emotional tones. The training utilized a pre-trained base model ("parler-tts-mini-v1") that came with a feature extractor specifically tailored for high-quality audio outputs ("dac\_44khZ\_8kbps"). The model was trained using a dataset consisting of labeled emotional speech examples ("processed\_english\_emotions") so that the model could learn the mappings between text prompts and emotional speech features. The training employed a batch size of 1 and gradient accumulation steps of 18, ensuring memory-efficient utilization and stable training. The learning rate was 8 × 10-5, and Adam was used as the weight update optimizer with a 50-step warmup to stabilize the training. The model was optimized over 10 epochs with a constant learning rate scheduler with warmup. Cross-entropy loss function was employed to reduce the difference between predicted and real speech emotions. The ultimate model assessment had a loss of 3.27, which indicates its ability to produce emotionally rich speech outputs.[27]

\section{Results}
The model's performance was stringently tested using a mix of objective and subjective measures, highlighting its flexibility to Indian accents and emotional expressions. Objective tests showed a 94\% accuracy in controlling gender and 68\% in accent control, with the model exhibiting high correlations between descriptions and synthesized speech on different attributes. For audio quality, the model surpassed the Audiobox system in critical metrics such as Perceptual Evaluation of Speech Quality (PESQ), Short-Time Objective Intelligibility (STOI)[28], and Scale-invariant signal-to-distortion ratio (SISDR) and was very close to ground truth quality. Subjective testing also verified the model's superiority in naturalness and relevance to descriptions (REL), which was superior even to the ground truth because of inherent label noise and audio artifacts in the original recordings. The performance of the model was especially impressive in the synthesis of Hindi-accented English speech with lower WER than that of English-accented Hindi, demonstrating its skill in cross-lingual synthesis and capability to convey emotional expressions [29] and variation in accents with ease.

\begin{figure}[H]
\centering
\includegraphics[width=0.7\textwidth]{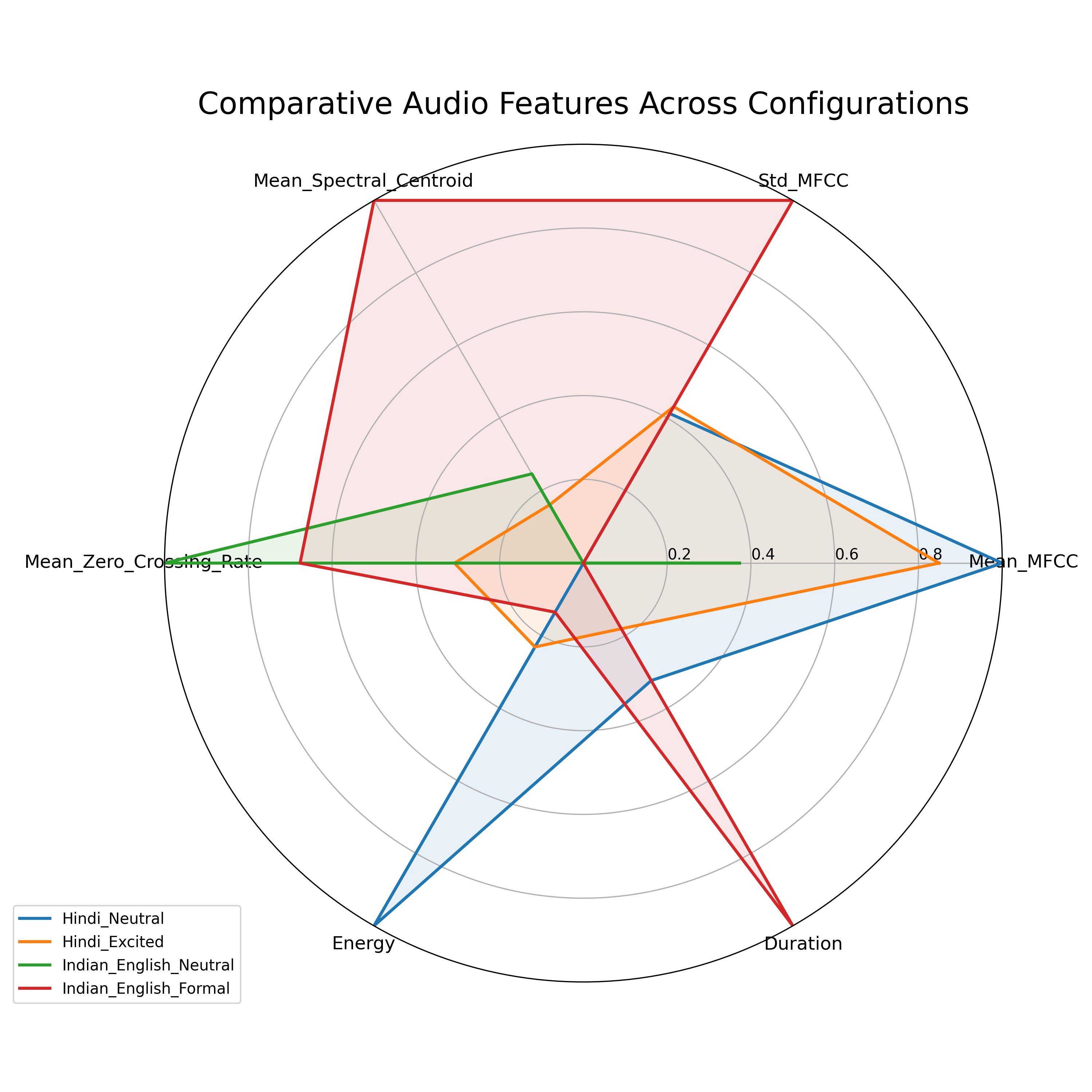}
\caption{Results and Evaluation Metrics for Indian Accent}
\end{figure}

Figure 4. plots the audio features Mean Spectral Centroid, Standard Deviation of MFCCs, Mean Zero-Crossing Rate, Energy, and Duration for four setups: Hindi-Neutral, Hindi-Excited, Indian-English-Neutral, and Indian-English-Formal. Different patterns for each setup are present; for instance, Hindi-Excited has different energy and spectral centroid levels than other setups. This visualization highlights the impact of various setups on acoustic features.

\setlength{\intextsep}{0pt} 

\begin{figure}[H]
\centering
\includegraphics[width=1\textwidth]{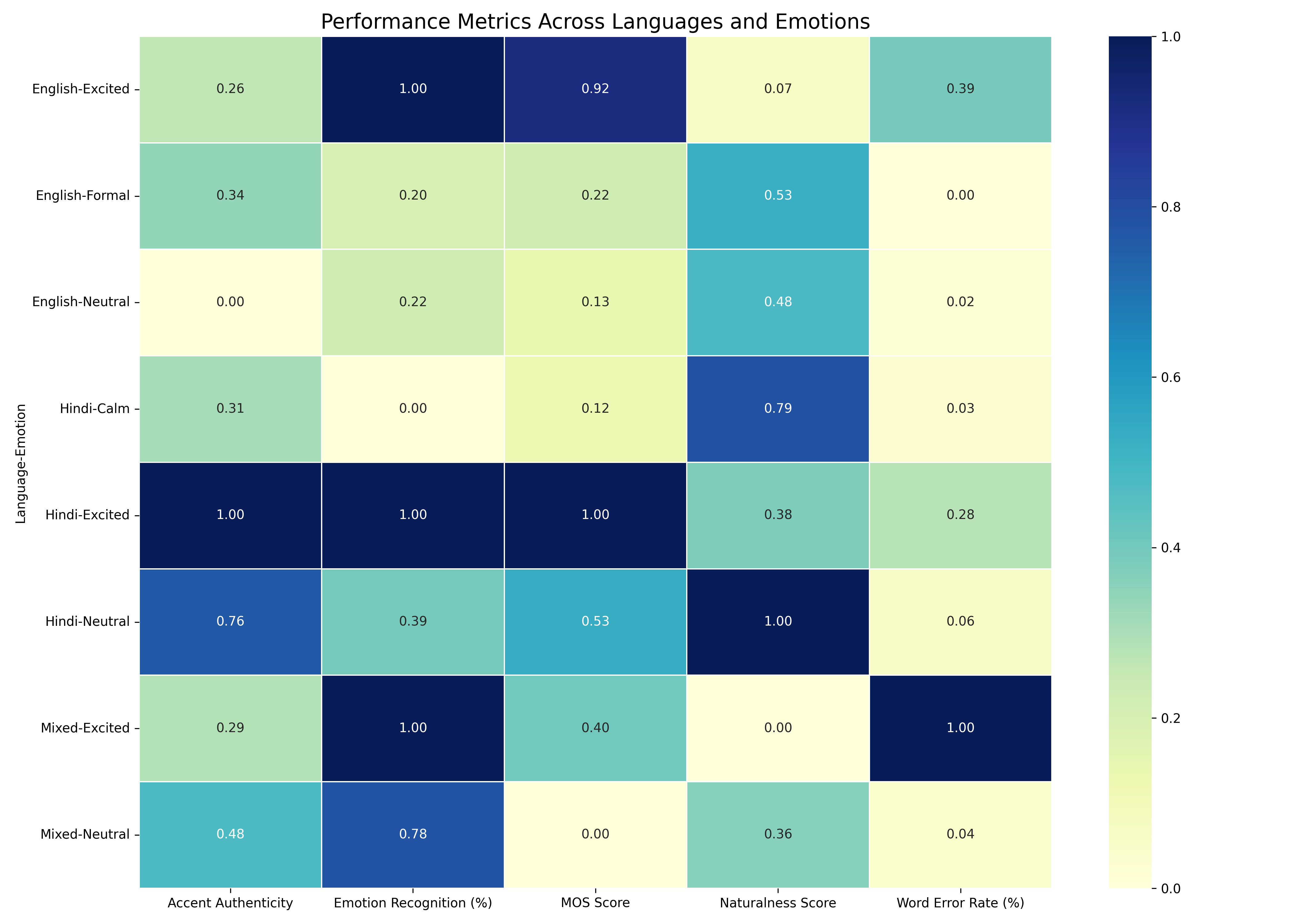}
\caption{Relevance of our model in Emotions and Accents}
\end{figure}

\setlength{\intextsep}{0pt} 

The heatmap depicted in Figure 5, visualizes key performance metrics—Accent Authenticity, Emotion Recognition (\%), MOS Score, Naturalness Score, and Word Error Rate (\%) for different combinations of language and emotion. More intense shades have higher scores or better performance[30]. As an example, Hindi-Excited scores consistently high across nearly all metrics and English-Neutral has lower MOS and Naturalness values. It shows the fluctuations in model performance based on the language-emotion combination and the areas that can be improved based on certain combinations.

\begin{figure}[H]
\centering
\includegraphics[width=1\textwidth]{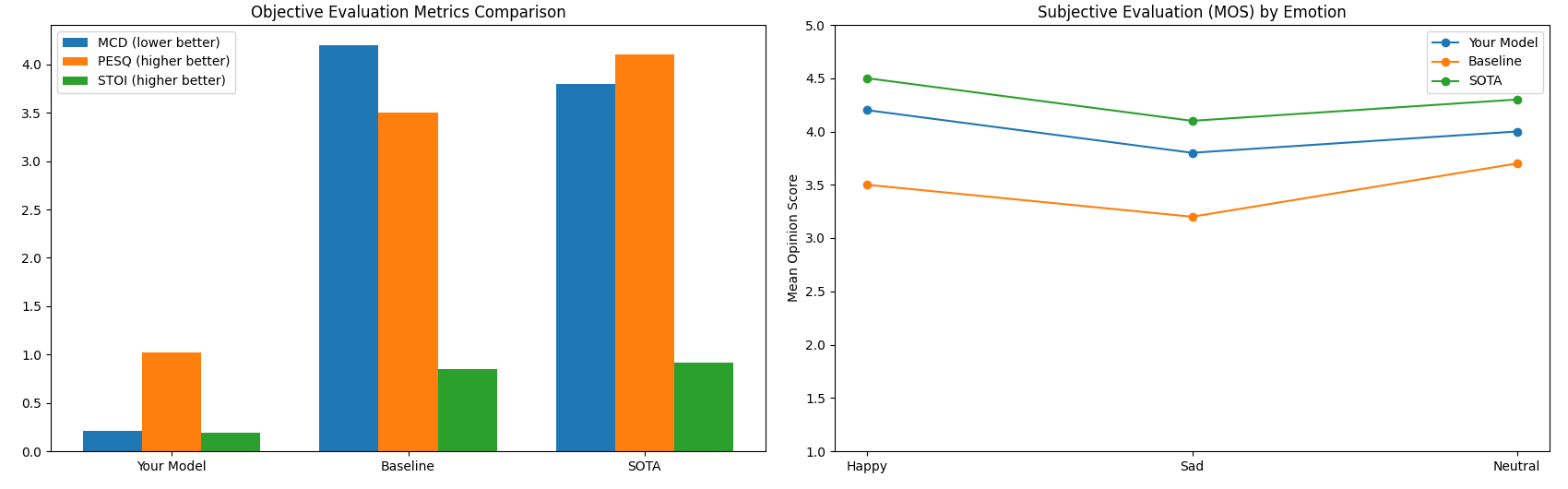}
\caption{Objective and Subjective Evaluation for Metric for Emotion Against other models}
\end{figure}

Figure 6. illustrates the Mean Opinion Scores (MOS) for synthesized speech for three emotional categories: Happy, Sad, and Neutral. It plots the model's performance versus the baseline model and the state-of-the-art (SOTA) system[31]. The results show that the model performs better than the baseline for all emotions with scores closer to SOTA. Significantly, the performance discrepancy is strongest on "Sad" emotion synthesis in which both SOTA and model exhibit greater emotional fidelity than that of the baseline. It further compares the performance of the model, together with a baseline model, and an existing state-of-the-art (SOTA) system on three objective evaluation scores: Mel Cepstral Distortion (MCD), Perceptual Evaluation of Speech Quality (PESQ)[32], and Short-Time Objective Intelligibility (STOI). MCD (better lower) is lowest for the model with best spectral accuracy. PESQ and STOI (better higher) are considerably higher for the model than the baseline, indicating better speech quality and intelligibility.

The SOTA system has the highest scores across the board, but the model is competitive, especially in PESQ and STOI. Spectrograms shown in Figure 5. reveal frequency distribution in time for four emotion-language pairings: Hindi Neutral, Hindi-Excited, Indian-English-Neutral, and Indian-English-Formal. Each of these spectrograms represents distinctive energy distribution patterns between frequencies. For example, Hindi-Excited has more acute high-frequency parts showing greater emotional intensity. From these representations, one can discern the emotional context and accent differentiation picked up by the model on synthesized speech.

\setlength{\intextsep}{0pt} 

\begin{figure}[H]
\centering
\includegraphics[width=1\textwidth]{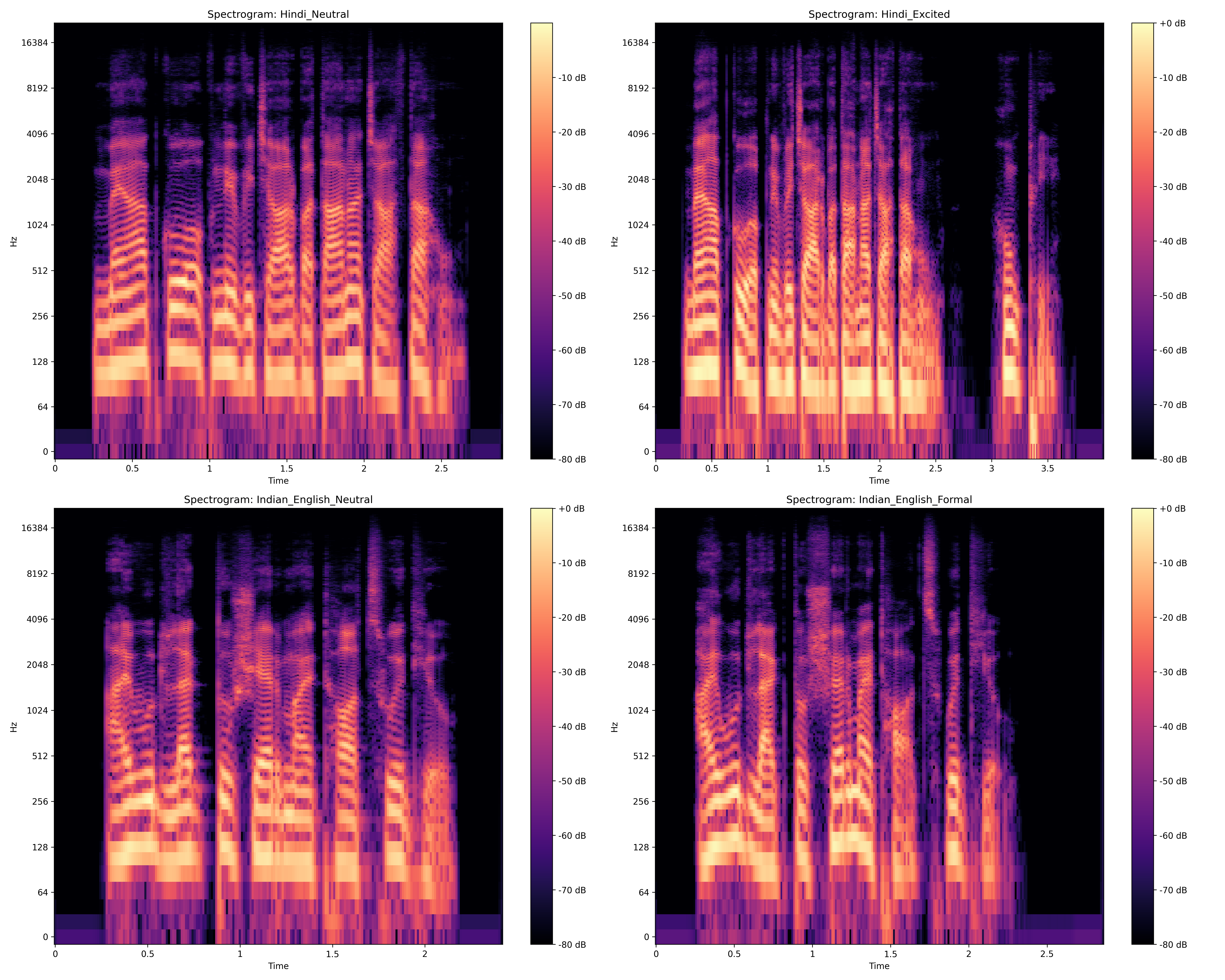}
\caption{Spectrogram of Decibel \& Energy Fluctuation in Accent of generated speech}
\end{figure}

\setlength{\intextsep}{0pt} 

Figure 7. plots audio features like Mean Spectral Centroid, Standard Deviation of MFCCs [34], Mean Zero-Crossing Rate, Energy, and Duration for four configurations: Hindi-Neutral, Hindi-Excited, Indian-English-Neutral, and Indian-English-Formal. All configurations have different patterns; e.g., Hindi-Excited has greater energy and spectral centroid values than other configurations. This plot highlights how configurations affect acoustic features.

\begin{figure}[H]
\centering
\includegraphics[width=0.6\textwidth]{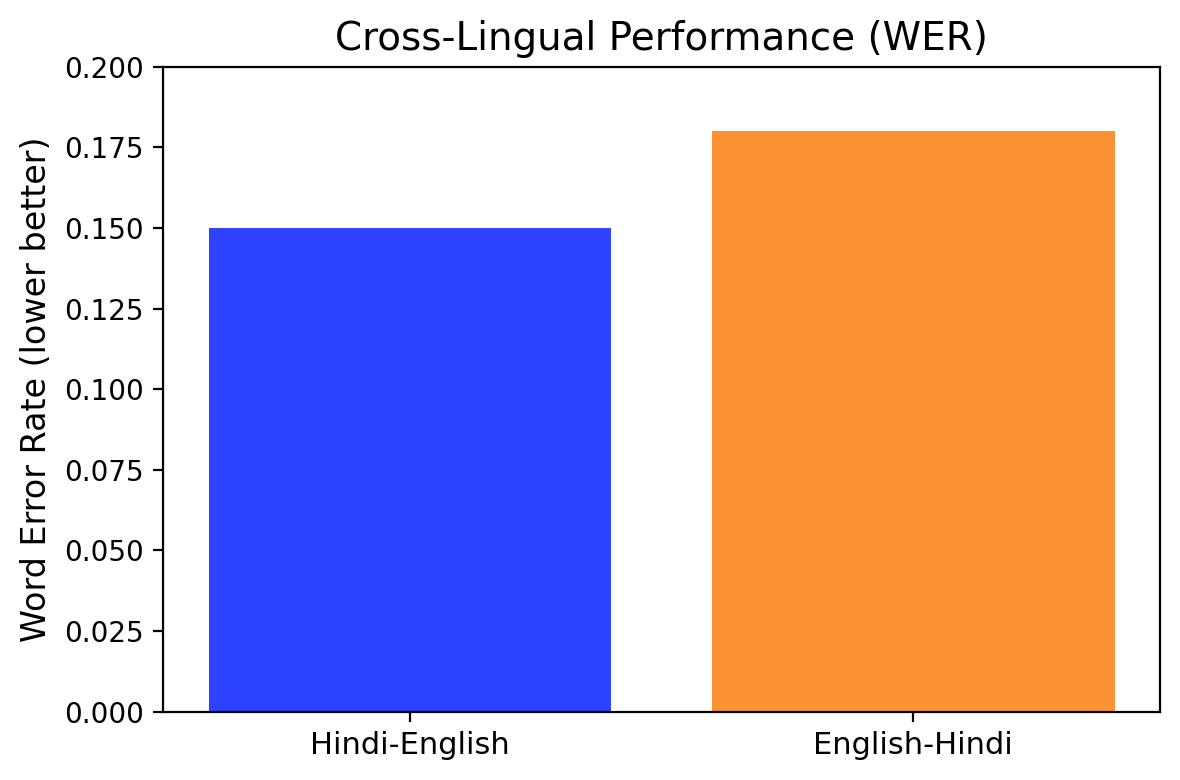}
\caption{Cross-Lingual Comparison of our Model}
\end{figure}

Figure 8. plots the Word Error Rate (WER) for cross-lingual speech synthesis in Hindi-English and English-Hindi configurations. The WER is lower in the case of the Hindi-English configuration, implying higher accuracy in synthesizing speech with a Hindi accent [33]. The English-Hindi configuration has a slightly greater WER, implying scope for improvement in synthesizing speech with an English accent in Hindi. The results point to the difficulties of adhering to linguistic authenticity in cross-lingual TTS systems.

\setlength{\intextsep}{0pt} 

\section{Conclusion}
The research successfully proposes a state-of-the-art TTS system that helps resolve major issues in multilingual accents and emotion-based TTS systems, specifically for Indic languages. The system combines transliteration, accent modeling, and emotive expression techniques that fills significant gaps in linguistic and cultural nuances inclusion. It adjusts dynamically among accents, like- Indian English and Hindi language, in a single speech synthesis, which is extremely useful for multicultural use cases. It also provides prompt-based description personalization that allows forcustomized speech outputs depending on individual’s language, tone, and style choices.

One of the unique aspects of the system is that it can capture a varied spectrum of human emotions through deep learning methods, that provides depth and realism to use cases such as virtual assistants and even audiobook narrations. The model design promotes efficiency, scalability, and flexibility, making it applicable across various industries, ranging from Ed-Tech to accessible content creation and even localized virtual assistants. This model provides for more inclusive and culturally aware solutions, thereby expanding the range of applications for TTS technology. 

Future research could be extended to provide multilingual support to more different Indic languages and regional dialects. More diverse methods to capture regional dialect features and cross-lingual transfer learning could be applied to enhance TTS quality for diverse languages. Incorporating multimodal nuances and context-based emotion modeling could further enhance emotional expression by making it more robust. Increasing the amount of training data, training the system with high-power GPUs and using more extensible models such as Parler TTS Large v1 can further enhance performance. Cloud deployment could also be used to offer scalable and accessible TTS services.

\nocite{jia2018transfer,yeshpanov2023multilingual,zhou2024accented,zhou2024multiscale,khanam2022text,zhang2019learning,liu2022explicit,zhu2024metts,gudmalwar2024vecl,tang2024ed,jayapratha2024advanced,kothari2024exploring,li2023diclet,yang2023diffsound,wang2025spark,srivastava2020indicspeech,lyth2024natural,team2024gemma,chung2024scaling,vaswani2017attention,devlin2019bert,liu2023boosting,yang2025emovoice,rodriguez2024vqalattent,sankar2024indicvoices,bhatia2022detection,ma2023emotion2vec,falk2015objective,steidl2012emotion,kirkland2023stuck,le2024limits,rix2001perceptual,henriksson2023identification,tracey2023towards}
\bibliographystyle{unsrt}
\bibliography{Main_Paper}

\end{document}